# SKIN TEXTURE RECOGNITION USING NEURAL NETWORKS

Nidhal K. Al abbadi[*], Nizar Saadi Dahir[*] & Zaid Abd Alkareem[*]

[*] IT Research and Development Center, Kufa University, Najaf - Iraq
nidhalka@it.kuiraq.com, nizarsd@it.kuiraq.com, zaidak@it.kuiraq.com

**ABSTRACT**

*Skin recognition is used in many applications ranging from algorithms for face detection, hand gesture analysis, and to objectionable image filtering. In this work a skin recognition system was developed and tested. While many skin segmentation algorithms relay on skin color, our work relies on both skin color and texture features (features derives from the GLCM) to give a better and more efficient recognition accuracy of skin textures. We used feed forward neural networks to classify input textures images to be skin or non skin textures. The system gave very encouraging results during the neural network generalization face.*

## 1. INTRODUCTION

Skin is a complex landscape that is difficult to model for many reasons. The skin texture features depends on many variables such as body location (knuckle vs. torso), subject parameters (age/gender/health) and imaging parameters (lighting and camera). Also as with many real world surfaces, skin appearance is strongly affected by the direction from which it is viewed and illuminated.

Recognition of human skin is an important task for both computer vision and graphics. For computer vision, accurate recognition of skin texture can greatly assist algorithms for human face recognition or facial feature tracking. In computer graphics, facial animation is an important problem which necessitates reliable skin texture recognition. In addition to computer vision and graphics, skin recognition is useful in dermatology and several industrial fields. In dermatology, the skin recognition can be used to develop methods for computer-assisted diagnosis of skin disorders, while in the pharmaceutical industry; quantification is useful when applied to measuring healing progress.

Many skin segmentation methods depend on skin color [5] [6] which have many difficulties. The skin color depends on human race and on lighting conditions, although this can be avoided in some ways using YCbCr color spaces in which the two components Cb and Cr depend only on chrominance, there still many problems with this method because there are many objects in the real world that have a chrominance in the range of the human skin which may be wrongly considered as skin. For the above reasons combining the texture features of skin with its color feature will increase the accuracy of skin recognition

## 2. RELATED WORK

Most existing skin segmentation techniques involve the classification of individual image pixels into skin and non-skin categories on the basis of pixel color. Numbers of comparative studies of skin color pixel classification have been reported. Jones and Rehg [4] created the first large skin database—the Compaq database—and used the Bayesian classifier with the histogram technique for skin detection. Brand and Mason [2] compared three different techniques on the Compaq database: thresholding the red/green ratio, color space mapping with 1D indicator, and RGB skin probability map. Terrillon et al. [9] compared Gaussian and Gaussian mixture models across nine chrominance spaces on a set of 110 images of 30 Asian and Caucasian people. Shin et al. [7] compared skin segmentation in eight color spaces. In their study, skin samples were taken from the Aleix Martinez and Robert Benavente (AR) and the University of Oulo face databases and non-skin samples were taken from the University of Washington image database.

## 3. FEATURE EXTRACTION
### 3.1. COLOR FEATURES (COLOR MOMENT)

Color moment is a compact representation of the color feature to characterize a color image. It has been shown that most of the color distribution information is captured by the three low-order moments. The first-order moment ($\mu_c$) captures the mean color, the second-order moment ($\sigma_c$) captures the standard deviation, and the third-order moment captures the skewness ($\theta_c$) of color. These three low-order moments ($\mu_c, \sigma_c, \theta_c$) are extracted for each of the three color planes (R G B), using the following mathematical formulation.

$$\mu_c = \frac{1}{MN}\sum_{i=1}^{M}\sum_{j=1}^{N} p_{ij}^c \qquad \ldots(1)$$

$$\sigma_c = \left[\frac{1}{MN}\sum_{i=1}^{M}\sum_{j=1}^{N}(p_{ij}^c - \mu_c)^2\right]^{1/2} \qquad \ldots(2)$$

$$\theta_c = \left[\frac{1}{MN}\sum_{i=1}^{M}\sum_{j=1}^{N}(p_{ij}^c - \mu_c)^3\right]^{1/3} \qquad \ldots(3)$$

Where M and N are the image dimensions, $p^c_{ij}$ is value of the $c^{th}$ color component of the color pixel in the $i^{th}$ row and $j^{th}$ column of the image. As a result, we need to extract only nine parameters (three moments for each of the three color planes) to characterize the color image [1].

### 3.2. TEXTURE FEATURES

Texture is a very interesting image feature that has been used for characterization of images, a major characteristic of texture is the repetition of a pattern or patterns over a region in an image. The elements of patterns are sometimes called textons. The size, shape, color, and orientation of the textons can vary over the region. The difference between two textures can be in the degree of variation of the textons. It can also be due to spatial statistical distribution of the textons in the image. Texture is an innate property of virtually all surfaces, such as bricks, fabrics, woods, papers, carpets, clouds, trees, lands and skin. It contains important information regarding underlying structural arrangement of the surfaces in an image.

Texture analysis has been an active area of research in pattern recognition. A variety of techniques have been used for measuring textural similarity. In 1973, Haralick et al. proposed co-occurrence matrix (GLCM) representation of texture features to mathematically represent gray level spatial dependence of texture in an image [1]. In this method the co-occurrence matrix is constructed based on the orientation and distance between image pixels. Meaningful statistics are extracted from this co-occurrence matrix, as the representation of texture. Since basic texture patterns are governed by periodic occurrence of certain gray levels, co-occurrence of gray levels at predefined relative positions can be a reasonable measure of the presence of texture and periodicity of the patterns. Several texture features such as entropy, energy, contrast, and homogeneity, can be extracted from the co-occurrence matrix of gray levels of an image.

The gray level co-occurrence matrix $C(i,j)$ is defined by first specifying a displacement vector $d_{x,y} = (\delta_x, \delta_y)$ (where $\delta x, \delta y$ are the displacements in the x and y directions respectively) and then counting all pairs of pixels separated by displacement $d_{x,y}$ and having gray levels $i$ and $j$. The matrix $C(i,j)$ is normalized by dividing each element in the matrix by the total number of pixel pairs.

Using this co-occurrence matrix, the texture features metrics are computed as follows.

$$Entropy = \sum_i \sum_j C(i,j) log(C(i,j)) \quad \ldots(4)$$

$$Energy = \sum_i \sum_j C^2(i,j) \quad \ldots(5)$$

$$Contrast = \sum_i \sum_j (i-j)^2 C(i,j) \quad \ldots(6)$$

$$Homogeneity = \sum_i \sum_j \frac{C(i,j)}{1+|i-j|} \quad \ldots(7)$$

These four features are combined with the nine features computed for each color component resulting in 13 element features vector use to characterize the skin texture in this work.

## 4. THE PROPOSED SKIN RECOGNITION ALGORITHM

Our proposed skin texture recognition algorithm consists of three main tasks as shown in figure 1:
(1) Creation of the library of representative skin features,
(2) Neural Network Training, and
(3) Classification.

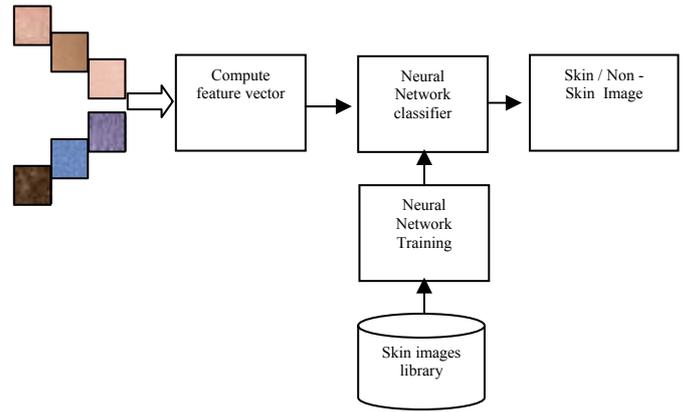

Figure 1: skin texture recognition algorithm tasks.

### 4.1. SKIN TEXTURE LIBRARY

The library of skin texture is comprised of 300 images of skin textures of size 80X80 pixel. The library consisted of variety of skin types of different human races, different places of the human body and different lighting conditions. Samples of that library is shown in figure 2.

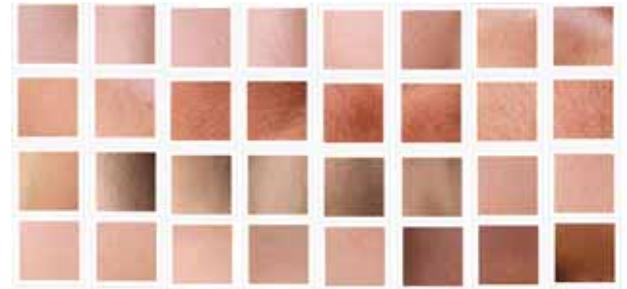

Figure 2: The skin library samples.

### 4.2 THE STRUCTURE OF THE NEURAL NETWORK

We used a feed forward backpropagation neural network with adaptable learning rate. The NN have 3 layer; an input layer (13 neuron), a hidden layer (50 neuron), and output layer (1 neuron). The activation function used is the tan sigmoid function, for both the hidden and the output layer. The input to the neural network is the feature vector containing 13 components these are the 4 texture feature and the three color moments for each color component (R G B), the NN has only one output as shown in figure 3.

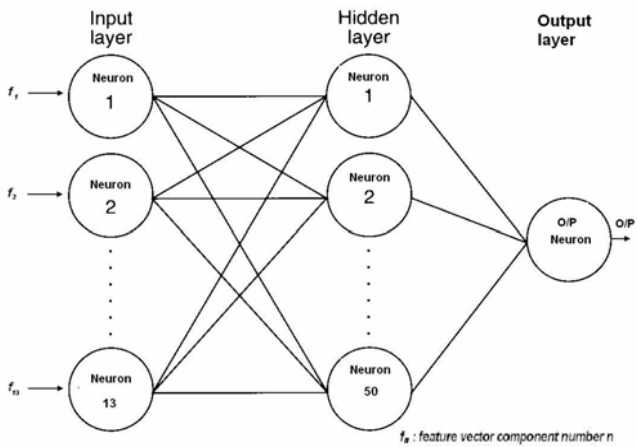

Figure 3: The neural network structure.

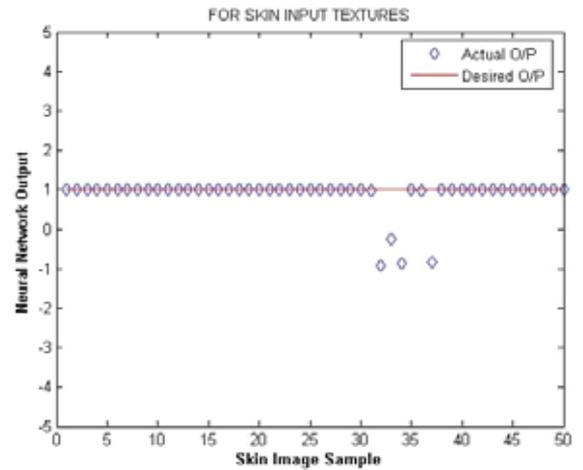

Figure 4: Neural network generalization O/P for skin input images

## 5. RESULTS

The NN training process is done using skin and non-skin texture images from the image library. The output of the neural network assumed to be 1 for skin input image and -1 for non-skin input image. The system where implemented using Matlab 7.0. The performance criterion used is SSE (sum square error) and the goal was $10^{-6}$ which is a very proved to be very acceptable goal. The performance goal is reached after 23039 training iteration. As shown in figure 3.

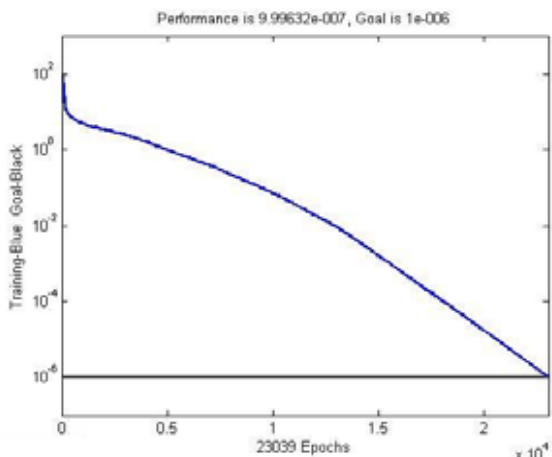

Figure 3: Neural Network Performance

To test our system we used 50 skin texture images and 50 non-skin texture images. These images were not used in the training phase and were taken under different lighting conditions, for different human races and form different perspectives. The images were input to the neural network after training to obtain the generalization results as shown in figures 4 and 5.

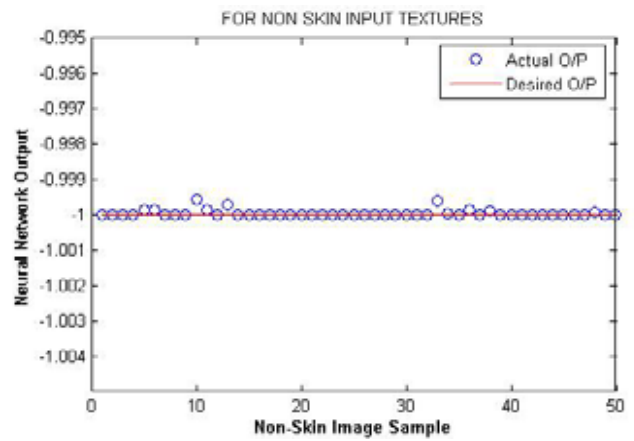

Figure 5: Neural Network Generalization O/P for non-skin input images

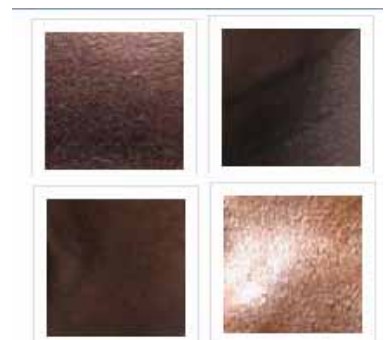

Figure 6: The undetected skin image samples.

## 6. DISCUSSION AND FUTURE WORK

It can be concluded from the figures bellow that the system gave very encouraging results for both skin and non-skin inputs. The use texture and color feature enhanced the performance of our system and gave recognition accuracy of 96% in the generalization test. This accuracy proves that the texture features are very useful as recognition features for skins in addition to color features that are used in many applications.

Figure 4 shows the output from the actual output form the neural network and the desired output which is 1 since the input images are all skin image. It can be seen that only four samples is falsely recognized as non-skin form 50 skin samples, these four samples are shown in figure 6. It can be seen from figure 6 that the images are either taken under lighting conditions that are very different from the lighting conditions under which that training set is taken, or they are not plane skin texture i.e. they contain 3D shadow. This shows the limitations of this method.

Figure 5 shows that all the non-skin image samples are truly detected as non skin (NN output of -1) with a very small error that is negligible.

The ultimate goal of this work is a system for objectionable image filtering. The future work is to develop algorithms for skin classification (classifying for which part of the body the skin belongs to), and to investigate for appropriate features that can serve for this purpose.

## REFERENCES

[1] Tinku Acharya "Image Processing Principles and Applications" A JOHN WILEY & SONS, MC. 2005.

[2] J. Brand and J. Mason, "A Comparative Assessment of Three Approaches to Pixel-Level Human Skin Detection," Proc. IEEE Int'l Conf. Pattern Recognition, vol. 1, pp. 1056-1059, Sept. 2000.

[3] R. M. Haralick, J. Shanmugam, and I. Dinstein, "Texture feature for image classification," *IEEE Transactions on Systems, Man, and Cybernetics,* 3, 610-621, 1973.

[4] M.J. Jones and J.M. Rehg, "Statistical Color Models with Application to Skin Detection," Int'l J. Computer Vision, vol. 46, no. 1, pp. 81-96, Jan. 2002.

[5] Hwei-Jen Lin, Shu-Yi Wang, Shwu-Huey, and Yang –Ta-Kao " Face Detection Based on Skin Color Segmentation and Neural Network" IEEE Transactions on, Volume: 2, pp1144- 1149, ISBN: 0-7803-9422-4, 2005.

[6] Son Lam Phung, Abdesselam Bouzerdoum, and Douglas Chai "Skin Segmentation Using Color And Edge Information" IEEE ISSPA ISBN: 0-7803-7946-2 2003.

[7] M.C. Shin, K.I. Chang, and L.V. Tsap, "Does Colorspace Transformation Make Any Difference on Skin Detection?" Proc. IEEE Workshop Applications of Computer Vision, pp. 275-279, Dec. 2002.

[8] F. Smach, et. al "Design of a Neural Network Classifier for Face Detection" Journal of Computer Science 257-260, ISSN 1549-3636, 2006

[9] J.-C. Terrillon, M.N. Shirazi, H. Fukamachi, and S. Akamatsu, "Comparative Performance of Different Skin Chrominance Models and Chrominance Spaces for the Automatic Detection of Human Faces in Color Images," Proc. IEEE Int'l Conf. Automatic Face and Gesture Recognition, pp. 54-61, Mar. 2000.